\documentclass[11pt]{article}

\newif \iffinal \finalfalse

\usepackage{amssymb}
\usepackage{amsfonts}
\usepackage{amsmath}
\usepackage{latexsym}
\usepackage{epsfig}
\usepackage{bm}
\usepackage{xspace}

\newenvironment{proof}{\noindent \textbf{Proof:}\nopagebreak[2]}{$\qed$}

\newcommand{\qed}{\hfill\rule{7pt}{7pt} \medskip}
\newtheorem{claim}{Claim}
\newtheorem{proposition}[claim]{Proposition}

\newtheorem{theorem}{Theorem}
\newtheorem{definition}{Definition}
\newtheorem{corollary}[claim]{Corollary}

\newtheorem{fact}[theorem]{Fact}

\newcommand{\ignore}[1]{}

\newcommand{\eps}{\epsilon}

\newcommand{\R}{\mathbf{R}}

\newcommand{\muij}{\mu^i_j}
\newcommand{\sigmaij}{\sigma^i_j}

\newcommand{\E}{{\bf E}}
\newcommand{\Var}{\mathrm{Var}}

\newcommand{\X}{{\bf X}}
\newcommand{\Z}{{\bf Z}}
\newcommand{\M}{{M}}

\newcommand{\cand}{(\{ \hat{\pi}^1,\dots,\hat{\pi}^k \}, \{\hat{\mu}_1^1, \hat{\mu}_2^1, \ldots,\hat{\mu}_n^k \})}

\newcommand{\alphamingauss}{\alpha_0}
\newcommand{\betamaxgauss}{\beta_0}
\newcommand{\etagauss}{\eta}

\newcommand{\epsall}{\eps_\mathrm{all}}
\newcommand{\epsI}{\eps_{\iset}}
\newcommand{\epsbig}{\eps_\mathrm{1}}
\newcommand{\epssmall}{\eps_\mathrm{2}}
\newcommand{\epsstar}{\eps_\mathrm{4}}
\newcommand{\epsprime}{\eps_\mathrm{3}}

\newcommand{\epsweights}{\eps_\mathrm{wts}}

\newcommand{\epsmeans}{\eps_\mathrm{means}}
\newcommand{\epsvars}{\eps_\mathrm{vars}}
\newcommand{\epsminweight}{\eps_\mathrm{minwt}}

\newcommand{\mumax}{\mu_\mathrm{max}}
\newcommand{\sigmamax}{\sigma_\mathrm{max}}
\newcommand{\sigmamin}{\sigma_\mathrm{min}}
\newcommand{\poly}{\mathrm{poly}}

\newcommand{\iset}{{\cal I}}
\newcommand{\mut}{\dot{\mu}}
\newcommand{\pit}{\dot{\pi}}
\newcommand{\pitt}{\ddot{\pi}}
\newcommand{\sit}{\dot{\sigma}}
\newcommand{\Zt}{\dot{\Z}}
\newcommand{\Xt}{\dot{\X}}

\newcommand{\W}{{\bf W}}

\newcommand{\bdd}{$(\mumax, \sigmamin^2,\sigmamax^2)$-bounded\xspace}
\newcommand{\ZML}{\Z^{\mathrm{ML}}}

\newcommand{\XX}{X}   

\newcommand{\PP}{{\bf P}}   
\newcommand{\QQ}{{\bf Q}}   
\newcommand{\KL}[2]   {\mathrm{KL}(#1 || #2)}   

\newcommand{\distset}{{\cal Q}}   
\newcommand{\lbound}{\alpha}                   
\newcommand{\ubound}{\beta}          
\newcommand{\hdist}{{{\QQ^*}}}   
\newcommand{\klbound}{\epsilon}  

\newcommand{\outdist}{{\QQ^{\mathrm{ML}}}}  

\newcommand{\samples}{{\cal S}}   
\newcommand{\dist}{\QQ}           

\newcommand{\mixalot}{{\sc WAM}\xspace}
\newcommand{\mixalotp}{{\sc WAM}$'$}

\renewcommand{\subsubsection}{\@startsection{subsubsection}{3}{0pt}{-12pt}{-5pt}{\normalsize\bf}}
\makeatother \iffinal
\newcommand{\fnote}[1]{}
\newcommand{\onote}[1]{}
\newcommand{\snote}[1]{}
\newcommand{\remove}[1]{}
\else
    \newcommand{\fnote}[1]{\footnote{{\bf [[Jon: {#1}\bf ]] }}}
    \newcommand{\snote}[1]{\footnote{{\bf [[Rocco: {#1}\bf ]] }}}
    \newcommand{\onote}[1]{\footnote{{\bf [[Ryan: {#1}\bf ]]}}}
    \newcommand{\remove}[1]{\par $<<${\it removed part}$>>$}
    \newcommand{\old}[1]{}
\fi

\begin{document}

\title{PAC Learning Mixtures of Axis-Aligned Gaussians with
No Separation Assumption}

\author{{Jon Feldman}\thanks{Some of this work was done while
supported by an NSF Mathematical Sciences Postdoctoral Research
Fellowship at Columbia University.}\\
Google\\
New York, NY\\
{\tt jonfeld@ieor.columbia.edu} \and
 \and {Ryan O'Donnell}\thanks{Some of
this work was done while at
the Institute for Advanced Study.}\\
Carnegie Mellon University\\
Pittsburgh, PA\\
{\tt ryanworldwide@gmail.com} \and
 {Rocco A.
Servedio}\thanks{Supported in part by NSF award CCF-0347282, by NSF
award CCF-0523664, and by a Sloan Foundation Fellowship.}\\
Columbia University\\
New York, NY\\
{\tt rocco@cs.columbia.edu} }

\maketitle

\begin{abstract}
We propose and analyze a new vantage point for the learning of
mixtures of Gaussians: namely, the PAC-style model of learning
probability distributions introduced by
Kearns~et~al.~\cite{KMR+:94short}.  Here the task is to construct a
hypothesis mixture of Gaussians that is statistically
indistinguishable from the actual mixture generating the data;
specifically, the KL~divergence should be at most $\eps$.

\quad In this scenario, we give a $\poly(n/\eps)$ time algorithm
that learns the class of mixtures of any constant number of
axis-aligned Gaussians in $\R^n$.  Our algorithm makes \emph{no}
assumptions about the separation between the means of the Gaussians,
nor does it have any dependence on the minimum mixing weight.  This
is in contrast to learning results known in the ``clustering''
model, where such assumptions are unavoidable.

\quad Our algorithm relies on the method of moments, and a
subalgorithm developed in~\cite{FOS:05focsshort} for a discrete
mixture-learning problem.

\end{abstract}



\section{Introduction}


In~\cite{KMR+:94short} Kearns et al.\ introduced an elegant and
natural model of learning unknown probability distributions. In this
framework we are given a class ${\cal C}$ of probability
distributions over $\R^n$ and access to random data sampled from an
unknown distribution $\Z$ that belongs to ${\cal C}.$ The goal is to
output a hypothesis distribution $\Z'$ which with high confidence is
$\eps$-close to $\Z$ as measured by the the Kullback-Leibler (KL)
divergence, a standard measure of the distance between probability
distributions (see Section~\ref{sec:prelim} for details on this
distance measure). The learning algorithm should run in time
$\poly(n/\eps)$.  This model is well-motivated by its close analogy
to Valiant's classical Probably Approximately Correct (PAC)
framework for learning Boolean functions~\cite{Valiant:84}.

Several notable results, both positive and negative, have been
obtained for learning in the Kearns et al.\ framework
of~\cite{KMR+:94short}, see, e.g., \cite{FKR+:97,Naor:96}.  Here we
briefly survey some of the positive results that have been obtained
for learning various types of \emph{mixture distributions}. (Recall
that given distributions $\X^1, \dots, \X^k$ and mixing weights
$\pi^1, \dots, \pi^k$ that sum to 1, a draw from the corresponding
mixture distribution is obtained by first selecting $i$ with
probability $\pi^i$ and then making a draw from $\X^i$.) Kearns et
al.\ gave an efficient algorithm for learning certain mixtures of
\emph{Hamming balls}; these are product distributions over
$\{0,1\}^n$ in which each coordinate mean is either $p$ or $1-p$ for
some $p$ fixed over all mixture components.  Subsequently Freund and
Mansour~\cite{FreundMansour:99short} and independently Cryan et
al.~\cite{CGG:02} gave efficient algorithms for learning a mixture
of two arbitrary product distributions over $\{0,1\}^n$. Recently,
Feldman et al.~\cite{FOS:05focsshort} gave a $\poly(n)$-time
algorithm that learns a mixture of any $k=O(1)$ many arbitrary
product distributions over the discrete domain $\{0,1,\dots,b-1\}^n$
for any $b=O(1)$.

\subsection{Results}

As described above, research on learning mixture distributions in
the PAC-style model of Kearns et al.\ has focused on distributions
over discrete domains. In this paper we consider the natural problem
of learning mixtures of Gaussians in the PAC-style framework
of~\cite{KMR+:94short}. Our main result is the following theorem:

\begin{theorem} \label{thm:maininformal}
{\bf (Informal version)} Fix any $k=O(1)$, and let $\Z$ be any
unknown mixture of axis-aligned Gaussians over $\R^n.$ There is an
algorithm that, given samples from $\Z$ and any $\eps$, $\delta > 0$
as inputs, runs in time $\poly(n/\eps)\cdot \log(1/\delta)$ and with
probability $1 - \delta$ outputs a mixture $\Z'$ of $k$ axis-aligned
Gaussians over $\R^n$ satisfying $KL(\Z||\Z') \leq \eps.$
\end{theorem}

A signal feature of this result is that it requires no assumptions
about the Gaussians being ``separated'' in space.  It also has no
dependence on the minimum mixing weight.  We compare our result with
other works on learning mixtures of Gaussians in the next section.

Our proof of Theorem~\ref{thm:maininformal} works by extending the
basic approach for learning mixtures of product distributions over
discrete domains from~\cite{FOS:05focsshort}.  The main technical
tool introduced in~\cite{FOS:05focsshort} is the \mixalot (Weights
And Means) algorithm; the correctness proof of \mixalot is based on
an intricate error analysis using ideas from the singular value
theory of matrices.  In this paper, we use this algorithm in a
continuous domain to estimate the parameters of the Gaussian
mixture.  Dealing with this more complex class of distributions
requires tackling a whole new set of issues around sampling error
that did not exist in the discrete case.

Our results strongly suggest that the techniques introduced
in~\cite{FOS:05focsshort} (and extended here) extend to PAC learning
mixtures of other classes of product distributions, both discrete
and continuous, such as exponential distributions or Poisson
distributions.  Though we have not explicitly worked out those
extensions in this paper, we briefly discuss general conditions
under which our techniques are applicable in
Section~\ref{sec:discussion}.

\subsection{Comparison with other frameworks for learning mixtures of Gaussians} \label{sec:clustercompare}

There is a vast literature in statistics on modeling with mixture
distributions, and on estimating the parameters of unknown such
distributions from data.  The case of mixtures of Gaussians is by
far the most studied case; see, e.g.,~\cite{Lindsay:95,TSM:85} for
surveys.  Statistical work on mixtures of Gaussians has mainly
focused on finding the distribution parameters (mixing weights,
means, and variances) of \emph{maximum likelihood}, given a set of
data.  Although one can write down equations whose solutions give
these maximum likelihood values, solving the equations appears to be
a computationally intractable problem.  In particular, the most
popular algorithm used for solving the equations, the \emph{EM
Algorithm} of Dempster~et~al.~\cite{DempsterLairdRubin:77short}, has
no efficiency guarantees and may run slowly or converge only to
local optima on some instances.

A change in perspective led to the first provably efficient
algorithm for learning: In 1999, Dasgupta~\cite{Dasgupta:99}
suggested learning in the \emph{clustering} framework.  In this
scenario, the learner's goal is to group all the sample points
according to which Gaussian in the mixture they came from.  This is
the strongest possible criterion for success one could demand; when
the learner succeeds, it can easily recover accurate approximations
of all parameters of the mixture distribution.  However, a strong
assumption is required to get such a strong outcome: it is clear
that the learner cannot possibly succeed unless the Gaussians are
guaranteed to be sufficiently ``separated'' in space.  Informally,
it must at least be the case that, with high probability, no sample
point ``looks like'' it might have come from a different Gaussian in
the mixture other than the one that actually generated it.

Dasgupta gave a polynomial time algorithm that could cluster a
mixture of \emph{spherical} Gaussians of \emph{equal radius}.  His
algorithm required separation on the order of $n^{1/2}$ times the
standard deviation. This was improved to $n^{1/4}$ by Dasgupta and
Schulman~\cite{DasguptaSchulman:00short}, and this in turn was
significantly generalized to the case of completely general (i.e.,
elliptical) Gaussians by Arora and Kannan~\cite{AroraKannan:01}.
Another breakthrough came from Vempala and
Wang~\cite{VempalaWang:02short} who showed how the separation could
be reduced, in the case of mixtures of $k$ spherical Gaussians (of
different radii), to the order of $k^{1/4}$ times the standard
deviation, times factors logarithmic in $n$.    This result was
extended to mixtures of general Gaussians (indeed, log-concave
distributions) in works by Kannan~et~al.~\cite{KSV:05short} and
Achlioptas and McSherry~\cite{AchlioptasMcSherry:05short}, with some
slightly worse separation requirements.  It should also be mentioned
that these results all have a running time dependence that is
polynomial in $1/\pi_{\text{min}}$, where $\pi_{\text{min}}$ denotes
the minimum mixing weight.

Our work gives another learning perspective that allows us to deal
with mixtures of Gaussians that satisfy \emph{no} separation
assumption. In this case clustering is simply not possible; for any
data set, there may be many different mixtures of Gaussians under
which the data are plausible. This possibility also leads to the
seeming intractability of finding the \emph{maximum} likelihood
mixture of Gaussians.  Nevertheless, we feel that this case is both
interesting and important, and that under these circumstances
identifying \emph{some} mixture of Gaussians which is statistically
indistinguishable from the true mixture is a worthy task. This is
precisely what the PAC-style learning scenario we work in requires,
and what our main algorithm efficiently achieves.

Reminding the reader that they work in significantly different
scenarios, we end this section with a comparison between other
aspects of our algorithm and algorithms in the clustering model. Our
algorithm works for mixtures of axis-aligned Gaussians.  This is
stronger than the case of spherical Gaussians considered
in~\cite{Dasgupta:99,DasguptaSchulman:00short,VempalaWang:02short},
but weaker than the case of general Gaussians handled
in~\cite{AroraKannan:01,KSV:05short,AchlioptasMcSherry:05short}.  On
the other hand, in Section~\ref{sec:discussion} we discuss the fact
that our methods should be readily adaptable to mixtures of a wide
variety of discrete and continuous distributions --- essentially,
any distribution where the ``method of moments'' from statistics
succeeds. The clustering algorithms discussed have polynomial
running time dependence on $k$, the number of mixture components,
whereas our algorithm's running time is polynomial in $n$ only if
$k$ is a constant. We note that in~\cite{FOS:05focsshort}, strong
evidence was given that (for the PAC-style learning problem that we
consider) such a dependence is unavoidable at least in the case of
learning mixtures of product distributions on the Boolean cube.
Finally, unlike the clustering algorithms mentioned, our algorithm
has no running time dependence on $1/\pi_{\text{min}}$.

\subsection{Overview of the approach and the paper}
\label{sec:outline}

An important ingredient of our approach is a slight extension of the
\mixalot algorithm, the main technical tool introduced
in~\cite{FOS:05focsshort}. The algorithm takes as input a parameter
$\eps
> 0$ and samples from an unknown mixture $\Z$ of $k$ product
distributions $\X^1,\dots,\X^k$ over $\R^n.$ The output of the
algorithm is a list of candidate descriptions of the $k$ mixing
weights and $kn$ coordinate means of the distributions
$\X^1,\dots,\X^k.$ Roughly speaking, the guarantee for the algorithm
proved in~\cite{FOS:05focsshort} is that with high probability at
least one of the candidate descriptions that the algorithm outputs
is ``good'' in the following sense:  it is an additive
$\eps$-accurate approximation to each of the $k$ true mixing weights
$\pi^1,\dots,\pi^k$ and to each of the true coordinate means
$\muij=\E[\X^i_j]$ for which the corresponding mixing weight $\pi^i$
is not too small. We give a precise specification  in
Section~\ref{sec:list}.

As described above, when \mixalot is run on a mixture distribution
it generates candidate estimates of mixing weights and means.
However, to describe a Gaussian we need not only its mean but also
its variance. To achieve this we run \mixalot \emph{twice}, once on
$\Z$ and once on what might be called ``$\Z^2$'' --- i.e., for the
second run, each time a draw $(z_1,\dots,z_n)$ is obtained from $\Z$
we convert it to $(z_1^2,\dots,z_n^2)$ and use that instead.  It is
easy to see that $\Z^2$ corresponds to a mixture of the
distributions $(\X^1)^2,\dots,(\X^k)^2$, and thus this second run
gives us estimates of the mixing weights (again) and also of the
coordinate \emph{second moments} $\E[(\X^i_j)^2]$. Having thus run
\mixalot twice, we essentially take the ``cross-product'' of the two
output lists to obtain a list of candidate descriptions, each of
which specifies mixing weights, means, and second moments of the
component Gaussians. In Section~\ref{sec:crossproduct} we give a
detailed description of this process and prove that with high
probability at least one of the resulting candidates is a ``good''
description (in the sense of the preceding paragraph) of the mixing
weights, coordinate means, and coordinate variances of the Gaussians
$\X^1,\dots,\X^k$.

To actually PAC learn the distribution $\Z$, we must find this good
description among the candidates in the list.  A natural idea is to
apply some sort of maximum likelihood procedure.  However, to make
this work, we need to guarantee that the list contains a
distribution that is close to the target in the sense of KL
divergence.  Thus, in Section~\ref{sec:hug}, we show how to convert
each ``parametric'' candidate description into a mixture of
Gaussians such that any additively accurate description indeed
becomes a mixture distribution with close KL divergence to the
unknown target. (This procedure also guarantees that the candidate
distributions satisfy some other technical conditions that are
needed by the maximum likelihood procedure.)  Finally, in
Section~\ref{sec:deathstroke} we put the pieces together and show
how a maximum likelihood procedure can be used to identify a
hypothesis mixture of Gaussians that has small KL divergence
relative to the target mixture.

%


\medskip

{\bf Note.}  This is the full version of ~\cite{FOS:06} which contains
all proofs omitted in that conference version because of space limitations.

\section{Preliminaries} \label{sec:prelim}


\noindent {\bf The PAC learning framework for probability
distributions.} We work in the Probably Approximately Correct model
of learning probability distributions which was proposed by Kearns
et al.~\cite{KMR+:94short}.  In this framework the learning
algorithm is given access to samples drawn from the target
distribution $\Z$ to be learned, and the learning algorithm must
(with high probability) output an accurate approximation $\Z'$ of
the target distribution $\Z$. Following~\cite{KMR+:94short}, we use
the {\em Kullback-Leibler (KL) divergence} (also known as the {\em
relative entropy}) as our notion of distance. The KL divergence
between distributions $\Z$ and $\Z'$ is
$$
\KL{\Z}{\Z'} := \int \Z(x) \ln (\Z(x)/\Z'(x))\,dx
$$
where here we have identified the distributions with their pdfs. The
reader is reminded that KL divergence is not symmetric and is thus
not a metric.  KL divergence is a stringent measure of the distance
between probability distances. In particular, it
holds~\cite{CoverThomas:91} that $ 0 \leq \|\Z - \Z'\|_2 \leq (2\ln
2) \sqrt{\KL{\Z}{\Z'}}$, where $\| \cdot \|_1$ denotes total
variation distance; hence if the KL divergence is small then so is
the total variation distance.

We make the following formal definition:
\begin{definition} \label{def:learn}
Let ${\cal D}$ be a class of probability distributions over $\R^n$.
An efficient (proper) learning algorithm for ${\cal D}$ is an
algorithm which, given $\epsilon$, $\delta > 0$ and samples drawn
from any distribution $\Z \in {\cal D}$, runs in
$\poly(n,1/\epsilon,1/\delta)$ time and, with probability at least
$1-\delta$, outputs a representation of a distribution $\Z' \in
{\cal D}$ such that $\KL{\Z}{\Z'} \leq \eps$.
\end{definition}

\noindent {\bf Mixtures of axis-aligned Gaussians.} Here we recall
some basic definitions and establish useful notational conventions
for later.

 A Gaussian distribution over $\R$ with mean $\mu$ and
variance $\sigma$ has probability density function $f(x) =
(1/\sqrt{2 \pi} \sigma) \exp\left( - {\frac {(x-\mu)^2}{2
\sigma^2}}\right).$ An \emph{axis-aligned} Gaussian over $\R^n$ is a
product distribution over $n$ univariate Gaussians.

\ignore{Throughout the remainder of the paper we write $\Z$ to
denote an unknown target mixture of $n$-dimensional axis-aligned
Gaussians $\X^1,\dots,\X^k$ with mixing weights
$\pi^1,\dots,\pi^k.$}

If we expect to learn a mixture of Gaussians, we need each Gaussian
to have reasonable parameters in each of its coordinates. Indeed,
consider just the problem of learning the parameters of a single
one-dimensional Gaussian: If the variance is enormous, we could not
expect to estimate the mean efficiently; or, if the variance was
extremely close to 0, any slight error in the hypothesis would lead
to a severe penalty in KL divergence.  These issues motivate the
following definition:

\begin{definition}
We say that $\X$ is a {\em  $d$-dimensional \bdd Gaussian} if $\X$
is a $d$-dimensional axis-aligned Gaussian with the property that
each of its one-dimensional coordinate Gaussians $\X_j$ has mean
$\mu_j \in [-\mumax, \mumax]$ and variance $(\sigma_j)^2 \in
[\sigmamin^2, \sigmamax^2].$
\end{definition}

\noindent {\bf Notational convention:} \emph{Throughout the rest of
the paper all Gaussians we consider are \bdd, where for notational
convenience we assume that the numbers $\mumax$, $\sigmamax^2$ are
at least 1 and that the number $\sigmamin^2$ is at most $1$. We will
denote by $L$ the quantity $\mumax \sigmamax/\sigmamin$, which in
some sense measures the bit-complexity of the problem. Given
distributions $\X^1,\dots,\X^k$ over $\R^n,$ we write $\mu^i_j$ to
denote $\E[\X^i_j]$, the $j$-th coordinate mean of the $i$-th
component distribution, and we write $(\sigmaij)^2$ to denote
$\Var[\X^i_j]$, the variance in coordinate $j$ of the $i$-th
distribution.}

\medskip

A mixture of $k$ axis-aligned Gaussians $\Z = \pi_1 \X^1 + \cdots +
\pi_k \X^k$ is completely specified by the parameters $\pi^i$,
$\muij,$ and $(\sigmaij)^2$.  Our learning algorithm for Gaussians
will have a running time that depends polynomially on $L$; thus the
algorithm is not strongly polynomial.



\section{Listing candidate weights and
means with \mixalot} \label{sec:list}


We first recall the basic features of the \mixalot algorithm
from~\cite{FOS:05focsshort} and then explain the extension we
require. The algorithm described in~\cite{FOS:05focsshort} takes as
input a parameter $\eps > 0$ and samples from an unknown mixture
$\Z$ of $k$ distributions $\X^1,\dots,\X^k$ where each $\X^i =
(\X^i_1, \dots, \X^i_n)$ is assumed to be a product distribution
over the bounded domain $[-1,1]^n$. \ignore{(Note that the results
of this section do not require that each $\X^i$ be a Gaussian
distribution; this will be important in
Section~\ref{sec:crossproduct}.)} The goal of \mixalot is to output
accurate estimates for the mixing weights $\pi^i$ and coordinate
means $\mu^i_j$; what the algorithm actually outputs is a list of
candidate ``parametric descriptions'' of the means and mixing
weights, where each candidate description is of the form $\cand$.

We now explain the notion of a ``good'' estimate of parameters from
Section~\ref{sec:outline} in more detail.  As motivation, note that
if a mixing weight $\pi^i$ is very low then the \mixalot algorithm
(or indeed any algorithm that only draws a limited number of samples
from $\Z$) may not receive any samples from $\X^i$, and thus we
would not expect \mixalot to construct an accurate estimate for the
coordinate means $\mu^i_1,\dots,\mu^i_n.$  We thus have the
following definition from~\cite{FOS:05focsshort}:

\begin{definition}  \label{def:paramacc}
A candidate $\cand$ is said to be \emph{parametrically
$\eps$-accurate} if:
\begin{enumerate}
 \item $|\hat{\pi}^i - \pi^i| \leq \eps$ for all $1 \leq i
 \leq k$;
 \item $|\hat{\mu}^i_j - \muij| \leq \eps$ for all $1 \leq i
 \leq k$ and $1 \leq j \leq n$ such that $\pi^i \geq \eps$.
\end{enumerate}
\end{definition}

Very roughly speaking, the~\mixalot algorithm
in~\cite{FOS:05focsshort} works by exhaustively ``guessing'' (to a
certain prescribed granularity that depends on $\eps$) values for
the mixing weights and for $k^2$ of the $kn$ coordinate means. Given
a guess, the algorithm tries to approximately solve for the
remaining $k(n-k)$ coordinate means using the guessed values and the
sample data; in the course of doing this the algorithm uses
estimates of the expectations $\E[\Z_j \Z_{j'}]$ that are obtained
from the sample data. From each guess the algorithm thus obtains one
of the candidates in the list that it ultimately outputs.

\ignore{Finally we describe the sense in which we require a slight
extension of~\mixalot.} The assumption  \cite{FOS:05focsshort} that
each distribution $\X^i$ in the mixture is over $[-1,1]^n$ has two
nice consequences:  each coordinate mean need only be guessed within
a bounded domain $[-1,1],$ and estimating $\E[\Z_j \Z_{j'}]$ is easy
for a mixture $\Z$ of such distributions. Inspection of the proof of
correctness of the \mixalot algorithm shows that these two
conditions are all that is really required. We thus introduce the
following:

\begin{definition}
Let $\X$ be a distribution over $\R$.  We say that $\X$ is {\em
$\lambda(\eps,\delta)$-samplable} if there is an algorithm ${\cal
A}$ which, given access to draws from $\X$, runs for
$\lambda(\eps,\delta)$ steps and outputs (with probability at least
$1 - \delta$ over the draws from $\X$) a quantity $\hat{\mu}$
satisfying $|\hat{\mu} - \E[\X]| \leq \eps$.
\end{definition}

With this definition in hand an obvious (slight) generalization of
\mixalot, which we denote \mixalotp, suggests itself.  The main
result about \mixalotp~that we need is  the following (the proof is
essentially identical to the proof in \cite{FOS:05focsshort} so we
omit it):

\begin{theorem} \label{thm:ourmain}
Let $\Z$ be a mixture of product distributions $\X^1,\dots,\X^k$
with mixing weights $\pi^1,\dots,\pi^k$ where each $\muij =
\E[\X^i_j]$ satisfies $|\muij| \leq U$ and $\Z_j \Z_{j'}$ is
$\poly(U/\eps)\cdot\log(1/\delta)$-samplable for all $j \neq j'$.
Given $U$ and any $\eps, \delta
> 0$, \mixalotp~runs in time $(n U/\eps)^{O(k^3)} \cdot \log(1/\delta)$ and outputs a list of $(n U/\eps)^{O(k^3)}$ many
candidates descriptions, at least one of which (with probability at
least $1-\delta$) is parametrically $\eps$-accurate.
\end{theorem}



\section{Listing candidate weights, means, and variances}
\label{sec:crossproduct}


Through the rest of the paper we assume that $\Z$ is a $k$-wise
mixture of independent \bdd Gaussians $\X^1,\dots,\X^k$, as
discussed in Section~\ref{sec:prelim}.  Recall also the notation $L$
from that section.

 As described in
Section~\ref{sec:outline}, we will run \mixalotp~twice, once on the
original mixture of Gaussians $\Z$ and once on the squared mixture
$\Z^2.$ In order to do this, we must show that both $\Z = \pi_1 \X^1
+ \cdots + \pi_k \X^k$ and $\Z^2 = \pi_1 (\X^1)^2 + \cdots + \pi_k
(\X^k)^2$ satisfy the conditions of Theorem~\ref{thm:ourmain}. The
bound $|\muij| \leq \mumax$ on coordinate means is satisfied by
assumption for $\Z$, and for $\Z^2$ we have that each
$\E[(\X^i_j)^2]$ is at most $\sigmamax^2 + \mumax^2$.  It remains to
verify the required samplability condition on products of two
coordinates for both $\Z$ and $\Z^2$; i.e.\ we must show that both
the random variables $\Z_j \Z_{j'}$ are samplable and that the
random variables $\Z_j^2\Z_{j'}^2$ are samplable. We do this in the
following proposition, whose straightforward but technical proof
appears in Appendix~\ref{sec:samplability}:

\begin{proposition}
\label{prop:sampprop} Suppose $\Z = (\Z_1,\Z_2)$ is the mixture of
$k$ two-dimensional \bdd  Gaussians.  Then both the random
 variable $\W := \Z_1\Z_2$ and the random variable $\W^2$
 are $\poly(L/\eps)\cdot\log(1/\delta)$-samplable.
\end{proposition}

The proof of the following theorem explains precisely how we can run
\mixalotp\ twice and how we can combine the two resulting lists (one
containing candidate descriptions consisting of mixing weights and
coordinate means, the other containing candidate descriptions
consisting of mixing weights and coordinate second moments) to
obtain a single list of candidate descriptions consisting of mixing
weights, coordinate means, and coordinate variances.

\begin{theorem}  \label{thm:cross}
Let $\Z$ be a mixture of $k=O(1)$ axis-aligned Gaussians
$\X^1,\dots,\X^k$ over $\R^n$, described by parameters $(
\{{\pi}^i\}, \{{\mu}^i_j\}, \{{\sigma}^i_j\})$. There is an
algorithm with the following property: For any $\eps$, $\delta > 0$,
given samples from $\Z$ the algorithm runs in $\poly(nL/\eps)\cdot
\log(1/\delta)$ time and with probability $1 - \delta$ outputs a
list of $\poly(nL/\eps)$ many candidates $( \{\hat{\pi}^i\},
\{\hat{\mu}^i_j\}, \{\hat{\sigma}^i_j\})$ such that for at least one
candidate in the list, the following holds:
\begin{enumerate}
\item $|\hat{\pi}^i - \pi^i| \leq \eps$ for all $i \in [k]$;
and
\item $|\hat{\mu}^i_j - \muij| \leq \eps$ and $|(\hat{\sigma}^i_j)^2 -
(\sigmaij)^2| \leq \eps$ for all $i,j$ such that $\pi^i \geq \eps$.
\end{enumerate}
\end{theorem}

\begin{proof}
First run the algorithm \mixalotp~with the random variable $\Z$,
taking the parameter ``$U$'' in \mixalotp~to be $L$, taking
``$\delta$'' to be $\delta/2$, and taking ``$\eps$'' to be $\eps/(6
\mumax).$  By Proposition~\ref{prop:sampprop} and
Theorem~\ref{thm:ourmain}, this takes at most the claimed running
time. \mixalotp~outputs a list List1 of candidate descriptions for
the mixing weights and expectations, List1 $= [\dots,
(\hat{\pi}^i,\hat{\mu}^i_j), \dots]$, which with probability at
least $1-\delta/2$ contains at least one candidate description which
is parametrically $\eps/(6 \mumax)$-accurate.

Define $(s^i_j)^2 = \E[(\X^i_j)^2] = (\sigmaij)^2 + (\muij)^2$. Run
the algorithm \mixalotp\ again on the squared random variable
$\Z^2$, with ``$U$'' $= \sigmamax^2 + \mumax^2$, ``$\delta$'' $=
\delta/2$, and ``$\eps$'' $= \eps/2.$  By
Proposition~\ref{prop:sampprop}, this again takes at most the
claimed running time.  This time \mixalotp\ outputs a list List2 of
candidates for the mixing weights (again) and second moments, List2
$= [\dots, (\hat{\hat{\pi}}^i,(\hat{s}^i_j)^2) \dots]$, which with
probability at least $1 - \delta/2$ has a ``good'' entry which
satisfies
\begin{enumerate}
\item $|\hat{\hat{\pi}}^i - \pi^i| \leq \eps/2$ for all $i =
1 \dots k$; and \item $|(\hat{s}^i_j)^2 - (s^i_j)^2| \leq \eps/2$
for all $i,j$ such that $\pi^i \geq \eps/2$.
\end{enumerate}

We now form the ``cross product'' of the two lists.  (Again, this
can be done in the claimed running time.)  Specifically, for each
pair consisting of a candidate $(\hat{\pi}^i,\hat{\mu}^i_j)$ in
List1 and a candidate $(\hat{\hat{\pi}}^i,(\hat{s}^i_j)^2)$ in
List2, we form a new candidate consisting of mixing weights, means,
and variances, namely $(\hat{\pi}^i, \hat{\mu}^i_j,
(\hat{\sigma}^i_j)^2)$ where $(\hat{\sigma}^i_j)^2 = (\hat{s}^i_j)^2
- (\hat{\mu}^i_j)^2.$ (Note that we simply discard
$\hat{\hat{\pi}}^i$.)

When the ``good'' candidate from List1 is matched with the ``good''
candidate from List2, the resulting candidate's mixing weights and
means satisfy the desired bounds.  For the variances, we have that
$|(\hat{\sigma}^i_j)^2 - (\sigmaij)^2|$ is at most
\[
|(\hat{s}^i_j)^2 - (s^i_j)^2| + |(\hat{\mu}^i_j)^2 - (\muij)^2| \leq
{\frac \eps 2} + |\hat{\mu}^i_j - \muij| \cdot |\hat{\mu}^i_j +
\muij| \leq {\frac \eps 2} + {\frac {\eps}{6 \mumax}} \cdot 3 \mumax
= \eps.\] This proves the theorem.
\end{proof}



\section{From parametric estimates to bona fide
distributions} \label{sec:hug}


At this point we have a list of candidate ``parametric''
descriptions $(\{\hat{\pi}^i\}, \{\hat{\mu}^i_j\},
\{(\hat{\sigma}^i_j)^2\})$ of mixtures of Gaussians, at least one of
which is parametrically accurate in the sense of
Theorem~\ref{thm:cross}.  In Section~\ref{sec:hug1} we describe an
efficient way to convert any parametric description into a true
mixture of Gaussians such that:

\begin{itemize}
\item [(i)] any
parametrically accurate description becomes a distribution with
close KL divergence to the target distribution; and
\item [(ii)] every mixture distribution that results from the conversion has a
pdf that satisfies certain upper and lower bounds (that will be
required for the maximum likelihood procedure).
\end{itemize}
The conversion procedure is conceptually straightforward --- it
essentially just truncates any extreme parameters to put them in a
``reasonable'' range --- but the details establishing correctness
are fairly technical.  By applying this conversion to each of the
parametric descriptions in our list from
Section~\ref{sec:crossproduct}, we obtain a list of mixture
distribution hypotheses all of which have bounded pdfs and at least
one of which is close to the target $\Z$ in KL divergence (see
Section~\ref{sec:convertlist}).  With such a list in hand, we will
be able to use maximum likelihood (in Section~\ref{sec:deathstroke})
to identify a single hypothesis which is close in KL divergence.

\subsection{The conversion procedure} \label{sec:hug1}

In this section we prove:

\begin{theorem} \label{thm:huggauss}
There is a simple efficient procedure ${\cal A}$ which takes values
\linebreak $(\{\hat{\pi}^i\}, \{\hat{\mu}^i_j\},
\{(\hat{\sigma}^i_j)^2\})$ and a value $\M>\mumax$ as inputs and
outputs a true mixture $\Zt$ of $k$ many $n$-dimensional \bdd
Gaussians with mixing weights $\pit^1, \dots, \pit^k$ satisfying
\begin{itemize}
\item [(a)] $\sum_{i=1}^k \pit^i = 1$, and

\item [(b)] $\alphamingauss \leq \Zt(x) \leq \betamaxgauss$ for all $x
\in [-\M,\M]^n$,
\end{itemize}
where $ \alphamingauss := \left[{\frac 1 {\sqrt{2 \pi} \sigmamax}}
\cdot \exp\left( {\frac {-2 \M^2} {\sigmamin^2}}\right)\right]^n$
and ~~$\betamaxgauss := 1/(\sqrt{2 \pi}\sigmamin)^n.$

Furthermore, suppose $\Z$ is a mixture of Gaussians
$\X^1,\dots,\X^k$ with mixing weights $\pi^i$, means $\muij$, and
variances $(\sigmaij)^2$ and that the following are satisfied:
\begin{itemize}
\item [(c)] for $i = 1 \dots k$ we have $|\pi^i - \hat{\pi}^i| \leq
\epsweights$ where $\epsweights \leq 1/(12k)^3$; and
\item [(d)] for all $i,j$ such that $\pi^i \geq
\epsminweight$ we have
 $|\muij - \hat{\mu}^i_j| \leq \epsmeans$ and $|(\sigmaij)^2 - (\hat{\sigma}^i_j)^2| \leq
 \epsvars$.
\end{itemize}
Then $\Zt$ will satisfy
$
\KL{\Z}{\Zt} \leq \etagauss(\epsmeans, \epsvars,
\epsweights,\epsminweight),
$
where
\begin{multline*}
\etagauss(\epsmeans, \epsvars, \epsweights,\epsminweight) := n \cdot
\left ( {\frac {\epsvars} {2 \sigmamin^2}} + {\frac {\epsmeans^2 +
\epsvars} {2(\sigmamin^2 - \epsvars)}} \right ) \\ + k \epsminweight
\cdot n \cdot \left( {\frac {\sigmamax^2 + 2
\mumax^2}{\sigmamin^2}}\right) + 13 k \epsweights^{1/3}.
\end{multline*}
\end{theorem}

\begin{proof}
We construct a mixture $\Zt$ of product distributions $\Xt^1, \dots,
\Xt^k$  by defining new mixing weights $\pit^i,$ expectations
$\mut^i_j,$ and variances $(\sit^i_j)^2.$ The procedure ${\cal A}$
is defined as follows:

\begin{enumerate}
\item
For all $i, j$, set
$$
{\mut}^i_j = \left \{
\begin{array}{lll}
-\mumax & \text{if $\hat{\mu}^i_j < -\mumax$}   \\
\mumax & \text{if $\hat{\mu}^i_j > \mumax$}   \\
\hat{\mu}^i_j & \text{o.w.}
\end{array}
\right. \quad \quad  \text{and} \quad \quad {\sit}^i_j = \left\{
\begin{array}{lll}
\sigmamin\ & \text{if $\hat{\sigma}^i_j < \sigmamin$}   \\
\sigmamax\ & \text{if $\hat{\sigma}^i_j > \sigmamax$}   \\
{\hat{\sigma}}^i_j & \text{o.w.}
\end{array}
\right.
$$

\item
For all $i = 1,\dots, k$ let $ \pitt^i = \left\{
\begin{array}{lll}
\hat{\pi}^i & \text{if $\hat{\pi}^i \geq \epsweights$}   \\
\epsweights & \text{if $\hat{\pi}^i < \epsweights$}.   \\
\end{array}
\right. $

Let $s$ be such that $s \sum_{i=1}^k \pitt^i = 1.$ Take $\pit^i = s
\pitt^i$. (This is just a normalization so the mixing weights sum to
precisely 1.)
\end{enumerate}

It is clear from this construction that condition (a) is satisfied.
For (b), the bounds on $\sit^i_j$ are easily seen to imply that
$\Xt^i(x) \leq 1/(\sqrt{2 \pi}\sigmamin)^n =: \betamaxgauss
$
for all $x \in \R^n$, and hence the same upper bound holds for the
mixture $\Zt(x)$, being a convex combination of the values
$\Xt^i(x)$.  Similarly, using the fact that $\M \geq \mumax$
together with the bounds on $\mut^i_j$ and $\sit^i_j$, we have that
$
 \Xt^i(x) \geq \left[{\frac 1 {\sqrt{2
\pi} \sigmamax}} \cdot \exp\left( {\frac {-2 \M^2}
{\sigmamin^2}}\right)\right]^n =: \alphamingauss,
$
for all $x \in[-\M,\M]^n,$ and this lower bound holds for $\Zt(x)$
as well.

We now prove the second half of the theorem; so suppose that
conditions (c) and (d) hold.  Our goal is to apply the following
proposition (proved in~\cite{FOS:05focsshort}) to bound
$\KL{\Z}{\Zt}$:

\begin{proposition} \label{prop:mixKL}
Let $\pi^1, \dots, \pi^k$, $\gamma^1, \dots, \gamma^k \geq 0$ be
mixing weights satisfying
$\sum \pi^i = \sum \gamma^i = 1$.
Let $\iset = \{ i : \pi^i \geq \epsprime \}$.
Let $\PP^1, \dots, \PP^k$ and $\QQ^1, \dots, \QQ^k$ be
distributions.
Suppose that
\begin{enumerate}
\item  $|\pi^i - \gamma^i| \leq \epsbig$
for all $i \in [k]$;
\item  $\gamma^i \geq \epssmall$ for all $i \in [k]$;
\item  $\KL{\PP^i}{\QQ^i} \leq \epsI$ for all $i \in \iset$;
\item  $\KL{\PP^i}{\QQ^i} \leq \epsall$ for all $i \in [k].$
\end{enumerate}
Then, letting $\PP$ denote the $\pi$-mixture of the $\PP^i$'s and
$\QQ$ the $\gamma$-mixture of the $\QQ^i$'s, for any
$\epsstar>\epsbig$ we have $ \KL{\PP}{\QQ} \leq \epsI + k \epsprime
\epsall  + k \epsstar \ln {\frac {\epsstar} {\epssmall}} + {\frac
\epsbig {\epsstar - \epsbig}}. \ignore{ {\frac k 2}
\epsweights^{1/2} \ln {\frac 1 {\epsweights}} + 2 \epsweights^{1/2}.
} $

\end{proposition}
More precisely, our goal is to apply this proposition with
parameters

\medskip
\begin{center}
$\epsbig = 3k\epsweights $; \; $\epssmall = \epsweights/2$; \;
$\epsprime = \epsminweight$; \; $\epsI = n \cdot \left ( {\frac
{\epsvars} {2 \sigmamin^2}} + {\frac {\epsmeans^2 + \epsvars}
{2(\sigmamin^2 - \epsvars)}} \right)$; \; $\epsall = n \cdot \left(
{\frac {\sigmamax^2 + 2 \mumax^2}{\sigmamin^2}}\right)$; \;
$\epsstar =\epsweights^{2/3}/2.$
\end{center}
To satisfy the conditions of the proposition, we must (1) upper
bound $|\pi^i - \pit^i|$ for all $i$; (2) lower bound $\pit^i$ for
all $i$; (3) upper bound $\KL{\X^i}{\Xt^i}$ for all $i$ such that
$\pi^i \geq \epsminweight$; and (4) upper bound $\KL{\X^i}{\Xt^i}$
for all $i.$ We now do this.

\medskip

\noindent {\bf (1) Upper bounding $|\pi^i - \pit^i|$.} A
straightforward argument given in~\cite{FOS:05focsshort} shows that
assuming $\epsweights \leq 1/(2k)$, we get  $|\pi^i - \pit^i| \leq 3
k \epsweights.$

\medskip

\noindent {\bf (2) Lower bounding $\pit^i$.}
In~\cite{FOS:05focsshort} it is also shown that $\pit^i \geq {\frac
\epsweights 2}$ assuming that $\epsweights \leq 1/k.$

\medskip

\noindent {\bf (3) Upper bounding $\KL{\X^i}{\Xt^i}$ for all $i$
such that $\pi^i \geq \epsminweight$.} Fix an $i$ such that $\pi^i
\geq \epsminweight$ and fix any $j \in [n].$ Consider some
particular $\muij$ and $\mut^i_j$ and $\sigmaij$ and $\sit^i_j$,  so
we have $|\muij - \hat{\mu}^i_j| \leq \epsmeans$ and $|(\sigmaij)^2
- (\hat{\sigma}^i_j)^2| \leq \epsvars$. Since $|\muij|\leq \mumax,$
by the definition of $\mut^i_j$ we have that $|\muij - \mut^i_j|
\leq \epsmeans$, and likewise we have $|(\sigmaij)^2 - (\sit^i_j)^2|
\leq \epsvars$. Let $\PP$ and $\QQ$ be the one-dimensional Gaussians
with means $\muij$ and $\mut^i_j$ and variances $\sigmaij$ and
$\sit^i_j$ respectively. By Corollary~\ref{cor:ubklgauss}, we have
\[
\KL{\PP}{\QQ} \leq {\frac {\epsvars} {2 \sigmamin^2}} + {\frac
{\epsmeans^2 + \epsvars} {2(\sigmamin^2 - \epsvars)}}.
\]
Each $\Xt^i$ is the product of $n$ such Gaussians. Since KL
divergence is additive for product distributions (see
Proposition~\ref{prop:nKL}) we have the following bound for each $i$
such that $\pi^i \geq \epsminweight$:
$$
\KL{\X^i}{\Xt^i} \leq n \cdot \left ( {\frac {\epsvars} {2
\sigmamin^2}} + {\frac {\epsmeans^2 + \epsvars} {2(\sigmamin^2 -
\epsvars)}} \right ).
$$

\noindent {\bf (4) Upper bounding $\KL{\X^i}{\Xt^i}$ for all $i \in
[k].$}  Using the fact that both $\X^i$ and $\Xt^i$ are \bdd, it
follows from Fact \ref{fact:klgauss} and Proposition \ref{prop:nKL}
that we have
$$ \KL{\X^i}{\Xt^i} \leq n \left( {\frac {\sigmamax^2 + 2
\mumax^2}{\sigmamin^2}}\right).
$$

\medskip

\noindent Proposition~\ref{prop:mixKL} now gives us
\[
\KL{\Z}{\Zt} \leq
n \cdot \left ( {\frac {\epsvars} {2 \sigmamin^2}} + {\frac
{\epsmeans^2 + \epsvars} {2(\sigmamin^2 - \epsvars)}} \right )
+ k \epsminweight \cdot n \cdot \left( {\frac {\sigmamax^2 + 2
\mumax^2}{\sigmamin^2}}\right)
+ R,
\]
where $R=k \epsstar \ln {\frac {\epsstar} {\epssmall}} + {\frac
\epsbig {\epsstar - \epsbig}} = {\frac k 2} \epsweights^{2/3} \ln
(\epsweights^{-1/3}) + {\frac {3 k \epsweights} {\epsweights^{2/3}/2
- 3 k \epsweights}}.$  Using the fact that $\ln x \leq x^{1/2}$ for
$x > 1$, the first of these two terms is at most ${\frac k 2}
\epsweights^{1/2}$. Using the fact that $\epsweights < 1/(12k)^3$,
the second of these terms is at most $12k\epsweights^{1/3}.$ So $R$
is at most $13 k \epsweights^{1/3}$ and the theorem is proved.
\end{proof}


\subsection{Getting a list of
distributions one of which is KL-close to the target}
\label{sec:convertlist}

In this section we show that combining the conversion procedure from
the previous subsection with the results of
Section~\ref{sec:crossproduct} lets us obtain the following:

\begin{theorem} \label{thm:somegoodKLgauss}
Let $\Z$ be any unknown mixture of $k = O(1)$ axis-aligned Gaussians
over $\R^n$.   There is an algorithm with the following property:
for any $\eps,\delta > 0$, given samples from $\Z$ the algorithm
runs in $\poly(nL/\eps)\cdot \log(1/\delta)$ time and with
probability $1 - \delta$ outputs a list of $\poly(nL/\eps)$ many
mixtures of Gaussians with the following properties:
\begin{enumerate}
\item For any $\M > \mumax$ such that $\M=\poly(nL/\eps)$,
every distribution $\Z'$ in the list satisfies
$\exp(-\poly(nL/\eps)) \leq \Z'(x) \leq \poly(L)^n$ for all $x \in
[-\M,\M]^n$.
\item Some distribution $\Z^\star$ in the
list satisfies $\KL{\Z}{\Z^\star} \leq  \eps$.
\end{enumerate}
\end{theorem}

Note that Theorem~\ref{thm:somegoodKLgauss} guarantees that $\Z'(x)$
has bounded mass only on the range $[-\M,\M]^n$, whereas the support
of $\Z$ goes beyond this range.  This issue is addressed in the
proof of Theorem~\ref{thm:learngaussians}, where we put together
Theorem~\ref{thm:somegoodKLgauss} and the maximum likelihood
procedure.

\medskip

\noindent {\bf Proof of Theorem~\ref{thm:somegoodKLgauss}:} We will
use a specialization of Theorem~\ref{thm:cross} in which we have
different parameters for the different roles that $\eps$ plays:

\medskip

\noindent {\bf Theorem~\ref{thm:cross}$'$}~\emph{ Let $\Z$ be a
mixture of $k=O(1)$ axis-aligned Gaussians $\X^1,\dots,\X^k$ over
$\R^n$, described by parameters $( \{{\pi}^i\}, \{{\mu}^i_j\},
\{{\sigma}^i_j\})$.  There is an algorithm with the following
property: for any $\epsmeans, \epsvars, \epsweights, \epsminweight,
\delta > 0$, given samples from $\Z$, with probability $1 - \delta$
it outputs a list of candidates $( \{\hat{\pi}^i\},
\{\hat{\mu}^i_j\}, \{\hat{\sigma}^i_j\})$ such that for at least one
candidate in the list, the following holds:
\begin{enumerate}
\item $|\hat{\pi}^i - \pi^i| \leq \epsweights$ for all $i \in [k]$;
and
\item $|\hat{\mu}^i_j - \muij| \leq \epsmeans$ and $|(\hat{\sigma}^i_j)^2 -
(\sigmaij)^2| \leq \epsvars$ for all $i,j$ such that $\pi^i \geq
\epsminweight$.
\end{enumerate}
The algorithm runs in time $\poly(nL/\eps')\cdot \log(1/\delta)$
where $\eps' = \min\{ \epsweights, \epsmeans, \epsvars,
\epsminweight\}.$}

\medskip

 Let $\eps, \delta > 0$ be given.  We run the algorithm of
Theorem~\ref{thm:cross}$'$ with parameters $\epsmeans = {\frac {\eps
\sigmamin^2}{12n}},$ $\epsvars = 2\epsmeans,$ $\epsminweight =
{\frac {\eps \sigmamin^2}{3kn(\sigmamax^2 + 2 \mumax^2)}}$ and
$\epsweights = {\frac {\eps^3}{(39k)^3}}.$  With these parameters
the algorithm runs in time poly$(nL/\eps) \cdot \log(1/\delta)$. By
Theorem~\ref{thm:cross}$'$, we get as output a list of
poly$(nL/\eps)$ many candidate parameter settings $(
\{\hat{\pi}^i\}, \{\hat{\mu}^i_j\}, \{\hat{\sigma}^i_j\})$ with the
guarantee that with probability $1-\delta$ at least one of the
settings satisfies
\begin{itemize}
\item $|\pi^i - \hat{\pi}^i| \leq \epsweights$ for all $i \in [k]$, and
\item $|\hat{\mu}^i_j - \muij| \leq \epsmeans$ and $|(\hat{\sigma}^i_j)^2 -
(\sigmaij)^2| \leq \epsvars$ for all $i,j$ such that $\pi^i \geq
\epsminweight$.
\end{itemize}

We now pass each of these candidate parameter settings through
Theorem~\ref{thm:huggauss}\ignore{with $\M$ chosen as specified in
Theorem~\ref{thm:somegoodKLgauss}}. (Note that $\epsweights <
1/(12k^3)$ as required by Theorem~\ref{thm:huggauss}.) By
Theorem~\ref{thm:huggauss}, for any $\M=\poly(nL/\eps)$ all the
resulting distributions will satisfy $\exp(-\poly(nL/\eps)) \leq
\Z'(x) \leq \poly(L)^n$ for all $x \in [-\M,\M]^n$. It is easy to
check that under our parameter settings, each of the three component
terms of $\etagauss$ (namely $n \cdot \left ( {\frac {\epsvars} {2
\sigmamin^2}} + {\frac {\epsmeans^2 + \epsvars} {2(\sigmamin^2 -
\epsvars)}} \right )$, $k \epsminweight \cdot n \left( {\frac
{\sigmamax^2 + 2 \mumax^2}{\sigmamin^2}}\right),$ and $13 k
\epsweights^{1/3}$) is at most $\eps/3$. Thus $\etagauss(\epsmeans,
\epsvars, \epsweights,\epsminweight) \leq \eps$, so at least one of
the resulting distributions $\Z^\star$ satisfies $\KL{\Z}{\Z^\star}
\leq \eps$.



\section{Putting it all together} \label{sec:deathstroke}


\subsection{Identifying a good distribution using maximum likelihood}

Theorem \ref{thm:somegoodKLgauss} gives us a list of distributions
at least one of which is close to the target distribution we are
trying to learn. Now we must {\em identify} some distribution in the
list which is close to the target.  We use a natural maximum
likelihood algorithm described in \cite{FOS:05focsshort} to help us
accomplish this:

\begin{theorem} \label{thm:needle} \cite{FOS:05focsshort}
Let $\ubound$, $\lbound$, $\klbound > 0$ be such that $\lbound <
\ubound.$
Let $\distset$ be a set of hypothesis distributions for some
distribution $\PP$ over the space $\XX$ such that at least one
$\hdist \in \distset$ has $\KL{\PP}{\hdist} \leq \klbound$. Suppose
also that $\lbound \leq \dist(x) \leq \ubound$ for all $\dist \in
\distset$ and all $x$ such that $\PP(x) > 0$.

Run the ML algorithm  on $\distset$ using a set $\samples$ of
independent samples from $\PP$, where $\samples = m.$
Then, with probability $1 - \delta$, where $\delta  \leq (|\distset|
+ 1) \cdot
           \exp \left ( {-2m \frac{\klbound^2}  {\log^2 \left ( \ubound / \lbound \right )} } \right ),
$
 the algorithm outputs some distribution $\outdist \in
 \distset$ which has $\KL{\PP}{\outdist} \leq 4\klbound$.
\end{theorem}

\subsection{The main result} \label{sec:mainresults}

Here we put the pieces together and give our main learning result
for mixtures of Gaussians.

\begin{theorem} \label{thm:learngaussians}
Let $\Z$ be any unknown mixture of $k$ $n$-dimensional Gaussians.
There is a $(nL/\epsilon)^{O(k^3)} \cdot \log(1/\delta)$  time
algorithm which, given samples from $\Z$ and any $\epsilon, \delta>
0$ as inputs, outputs a mixture $\Z'$ of $k$ Gaussians which with
probability at least $1 - \delta$ satisfies $\KL{\Z}{\Z'} \leq
\epsilon.$
\end{theorem}

\begin{proof}
Run the algorithm given by Theorem~\ref{thm:somegoodKLgauss}. With
probability $1-\delta$ this produces a list of $T =
(nL/\eps)^{O(k^3)} \cdot \log(1/\delta)$ hypothesis distributions,
one of which, $\Z^\star$, has KL divergence at most $\eps$ from $\Z$
and all of which have their pdfs bounded between
$\exp(-\poly(nL/\eps))$ and $\poly(L)^n$ for all $x \in [-\M,
\M]^n$, where $\M > \mumax$ is any $\poly(nL/\eps)$.

We now consider $\Z_\M,$ the $\M$-truncated version of $\Z$; this is
simply the distribution obtained by restricting the support of $\Z$
to be $[-\M,\M]^n$ and scaling so that $\Z_\M$ is a distribution
(see Appendix~\ref{ap:trunc} for a precise definition of $\Z_\M$).
We prove the following proposition in Appendix \ref{ap:trunc}:

\begin{proposition} \label{prop:trunc}
Let $\PP$ and $\QQ$ be any mixtures of $n$-dimensional  Gaussians.
Let $\PP_\M$ denote the $\M$-truncated version of $\PP$. For some
$\M=\poly(nL/\eps)$ we have $|\KL{\PP_M}{\QQ} - \KL{\PP}{\QQ}| \leq
4\eps + 2\eps \cdot \KL{\PP}{\QQ}$.
\end{proposition}
This proposition implies that $\KL{\Z_\M}{\Z^\star} \leq 7 \eps.$

Now run the ML algorithm with $m = \poly(nL/\eps)\log(\M/\delta)$ on
this list of hypothesis distributions \emph{using $\Z_\M$ as the
target distribution}. (We can obtain draws from $\Z_\M$ using
rejection sampling from $\Z$; with probability $1 - \delta$ this
incurs only a negligible increase in the time required to obtain $m$
draws.) Note that running the algorithm with $\Z_\M$ as the target
distribution lets us assert that all hypothesis distributions have
pdfs bounded above and below on the support of the target
distribution, as is required by Theorem \ref{thm:needle}. (In
contrast, since the support of $\Z$ is all of $\R^n$, we cannot
guarantee that our hypothesis distributions have pdf bounds on the
support of $\Z$.) By Theorem \ref{thm:needle}, with probability at
least $1-\delta$ the ML algorithm outputs a hypothesis $\ZML$ such
that $\KL{\Z_\M}{\ZML} \leq 28\eps.$

It remains only to bound $\KL{\Z}{\ZML}.$  By Proposition
\ref{prop:trunc} we have
$$
\KL{\Z}{\ZML} \leq 28 \eps + 4 \eps + 2 \eps\cdot\KL{\Z}{\ZML}
$$
which implies that $\KL{\Z}{\ZML} \leq 33 \eps.$ The running time of
the overall algorithm is $(nL/\epsilon)^{O(k^3)} \cdot
\log(1/\delta)$ and the theorem is proved.
\end{proof}


\section{Extensions to other distributions} \label{sec:discussion}

In this paper we have shown how to PAC learn mixtures of any
constant number of distributions, each of which is an
$n$-dimensional Gaussian product distribution.  This expands upon
the work by Feldman~et~al.~\cite{FOS:05focsshort} which worked for
discrete distributions in place of Gaussians.  It should be clear
from our work that in fact many ``nice'' univariate distributions
can be handled similarly.  Also, it should be noted that the $n$
coordinates need not come from the same family of distributions; for
example, our methods would handle mixtures where some attributes had
discrete distributions and the remainder had Gaussian distributions.

What level of ``niceness'' do our methods require  for a
parameterized family of univariate distributions on $\R$?  First and
foremost, it should be amenable to the ``method of moments'' from
statistics.  By this it is meant that it should be possible to solve
for the parameters of the distribution given a constant number of
the moments. Distributions in this category include gamma
distributions, chi-square distributions, beta distributions,
exponential --- more generally, Weibull --- distributions, and more.
As a trivial example, the unknown parameter of an exponential
distribution is simply its mean. As a slightly more involved
example, given a beta distribution with unknown parameters $\alpha$
and $\beta$ (the pdf for which is proportional to
$x^{\alpha-1}(1-x)^{\beta-1}$ on $[0,1]$), these parameters can be
determined from mean and variance estimates via
\[
\alpha = \E[\X]\left(\frac{\E[\X](1-\E[\X])}{\Var[\X]}-1\right),
\qquad \beta =
(1-\E[\X])\left(\frac{\E[\X](1-\E[\X])}{\Var[\X]}-1\right).
\]
So long as the univariate distribution family can be determined by a
constant number of moments, our basic strategy of running \mixalot
multiple times to determine moment estimates and then taking the
cross-products of these lists can be employed.

There are only two more concerns that need to be addressed for a
given parameterized family of distributions.  First, one needs an
analogue of Proposition~\ref{prop:sampprop}, showing that products
of independent random variables from the distribution family are
efficiently samplable.  (In fact, this should hold for
\emph{mixtures} of such, but this is very likely to be implied in
any reasonable case.)  This immediately holds for any distribution
with bounded support; it will also typically hold for ``reasonable''
probability distributions that have pdfs with rapidly decaying
tails.

Second, one needs an analogue of Theorem~\ref{thm:huggauss}.  This
requires that it should be possible to convert accurate candidate
parameter values into a KL-close actual distribution.  It seems that
this will typically be possible so long as the distributions in the
family are not highly concentrated at any particular point.  The
conversion procedure should also have the property that the
distributions it output have pdfs that are bounded below/above by at
most exponentially small/large values, at least on
polynomially-sized domains.  This again seems to be a mild
constraint, satisfiable for reasonable distributions with rapidly
decaying tails.

In summary, we believe that for most parameterized distribution
families ``$D$'' of interest, performing a small amount of technical
work should be sufficient to show that our methods can learn
``mixtures of products of $D$'s''.  We leave the problem of checking
these conditions for distribution families of interest as an avenue
for future research.

\appendix

\section{Notational convention on Gaussians}
\label{sec:notconvgauss}

Recall that all Gaussians we consider are \bdd.  In dealing with
Gaussians it will be very useful to define a function $\M(\theta)$
which satisfies
$$
\int_{|x| \geq \M} \X(x) dx < \theta, \int_{|x| \geq \M} |x|\X(x) dx
< \theta, \mbox{~~and~~} \int_{|x| \geq \M} x^2 \X(x) dx < \theta
$$
for any one-dimensional \bdd Gaussians $\X.$  Straightforward
arguments show that this can be achieved with
$\M(\theta)=\poly(L/\theta).$

\medskip

\noindent {\bf Notational convention:} \emph{Throughout the
appendices $\M(\theta)=\poly(L/\theta)$ denotes a function
satisfying the conditions above.}


\section{Proof of Proposition~\ref{prop:sampprop}}
\label{sec:samplability}


\begin{proof}
We shall prove the proposition for $\W^2$; the proof for $\W$ is
similar but slightly simpler.

Let the mixing weights be $\pi^1, \dots, \pi^k$ and suppose that
$\Z_j$ is a mixture of $\X_j^1, \dots, \X_j^k$ for $j = 1, 2$. Let
$s = \E[\W^2]$.

Recall the quantity $\M = \M(\theta)$ and take $C = \M^4 =
\poly(L/\theta)$.  Let $\W^2_C$ denote the random variable $\W^2$
conditioned on the event $|\W^2| \leq C$.  Observe that
\begin{equation} \label{eqn:suck}
\Pr[\W^2 > C] = \Pr[\W^2 > \M^4] \leq \Pr[|\Z_1| > \M] + \Pr[|\Z_2|
> \M] \leq 2\theta,
\end{equation}
using the fact that $\Z_1$ and $\Z_2$ are \bdd Gaussians and the
definition of $\M$.

We shall show that $|\E[\W^2_C] - s| \leq \eps/2$.  Our sampling
algorithm for $\W^2$ will be to sample from $\W^2_C$ using rejection
sampling and to compute and output the empirical mean of $\W^2_C$.
Since the random variable $\W^2_C$ is bounded in the range $[-C,
C]$, by the Hoeffding bound if we take $\poly(C/\eps)\cdot
\log(1/\delta) = \poly(L/\eps\theta)\cdot\log(1/\delta)$ samples
from $\W_C^2$ then with probability $1-\delta$ the empirical mean of
$\W_C^2$ will be within $\eps/2$ of the true mean $\E[\W_C^2]$.
(Technically, we must also note that since $\theta$ is much smaller
than $1$ we can do rejection sampling with very little slowdown.)
Thus it remains to show that indeed $|\E[(\W_C)^2] - s| \leq
\eps/2$.

Observe that $\E[(\W_C)^2] = \sum_{i=1}^k \pi^i \E[(\W_C)^2 \mid
\text{$i$ is chosen}]$ and $s = \sum_{i=1}^k \pi^i \E[\W^2 \mid
\text{$i$ is chosen}]$.  Thus by convexity it is sufficient to prove
$|\E[(\X_1^i)^2 (\X_2^i)^2 \mid (\X_1^i)^2(\X_1^i)^2 \leq C] -
\E[(\X_1^i)^2 (\X_2^i)^2]| \leq \eps/2$ for all $i = 1\dots k$. For
simplicity we now write $\X_j = \X_j^i$ for $j = 1, 2$. Recall that
$\X_1$ and $\X_2$ are one-dimensional \bdd Gaussians.

Let $p(w)$ be the pdf for the random variable $(\X_1)^2 (\X_2)^2$.
Note that
\begin{eqnarray}
\left|\int_{|w| > C} w p(w) dw\right| & = & \int_{x_1} \int_{x_2}
{\bf 1}_{\{x_1^2 x_2^2 \geq C\}} x_1^2x_2^2 \X_1(x_1)
\X_2(x_2) dx_1dx_2 \nonumber\\
& \leq & \int_{x_1} \int_{x_2} ({\bf 1}_{\{|x_1| \geq C^{1/4}\}} +
{\bf 1}_{\{|x_2| \geq C^{1/4}\}}) x_1^2x_2^2 \X_1(x_1)
\X_2(x_2) dx_1dx_2 \nonumber\\
& = & \int_{x_2} x_2^2 \X_2(x_2) dx_2 \int_{|x_1| \geq \M}
x_1^2 \X_1(x_1) dx_1  \nonumber\\
&& \qquad \qquad + \int_{x_1} x_1^2 \X_1(x_1) dx_1
\int_{|x_2| \geq \M} x_2^2 \X_2(x_2) dx_2 \nonumber\\
&=& \E[(\X_2)^2] \int_{|x_1| \geq \M} x_1^2 \X_1(x_1) dx_1\nonumber\\
&& \qquad \qquad + \E[(\X_1)^2]
\int_{|x_2| \geq \M} x_2^2 \X_2(x_2) dx_2 \nonumber\\
&\leq& 2L^2 \left(\int_{|x_1| \geq \M} x_1^2 \X_1(x_1) dx_1 +
\int_{|x_2| \geq \M} x_2^2 \X_2(x_2) dx_2\right) \nonumber\\
& \leq & 4\theta L^2,
\end{eqnarray}
using the definitions of $\M$ and $L$.

Let $\eta = 1/(1-\Pr[(\X_1)^2(\X_2)^2 > C]) - 1$, so $\eta \leq
3\theta$ using the same argument as in~(\ref{eqn:suck}).  Note that
the pdf $p_C(w)$ for the random variable $(\X_1)^2(\X_2)^2$
conditioned on $|(\X_1)^2(\X_2)^2| \leq C$ is given by
\[
p_C(w) = \left\{\begin{array}{ll} (1+\eta)p(w) & \text{if $|w| \leq
C$,}
\\ 0 & \text{if $|w| > C$.} \end{array} \right.
\]

\noindent Let $t = \E[(\X_1)^2 (\X_2)^2]$; finally, we can show that
$|\E[(\X_1)^2 (\X_2)^2 \mid (\X_1)^2(\X_2)^2 \leq C] - t| \leq
\eps/2$, as desired:
\begin{eqnarray*}
|\E[(\X_1)^2 (\X_2)^2 \mid (\X_1)^2(\X_2)^2 \leq C] - t|
& = & \left|\int_\R w p_C(w) - \int_\R w p(w)\right| \\
& = & \left|(1+\eta)\int_{|w| \leq C} wp(w) - \int_{|w| \leq C} wp(w) - \int_{|w| > C} wp(w)\right|\\
& = & \left|\eta\int_{|w| < C} wp(w) - \int_{|w| \geq C} wp(w) \right|\\
& \leq & \eta t + \theta \leq (3\theta)\poly(L) + \theta,
\end{eqnarray*}
once more using the definition of $\M$ (note: $C \geq \M$). Choosing
$\theta = \poly(\epsilon/L)$, we get that this is bounded by
$\eps/2$; consequently $\M = \poly(L/\epsilon)$ and the sampling
time is as claimed.
\end{proof}


\section{Auxiliary facts about KL divergence}
\label{sec:auxiliary}

The following fact gives the KL divergence  between two univariate
Gaussians; it can be found in, e.g.,~\cite{Seeger:03}.

\begin{fact} \label{fact:klgauss}
Let $\PP, \QQ$ each be a one-dimensional normal distribution with
means and variances  $\mu_{\PP}, \sigma_{\PP}$ and $\mu_{\QQ},
\sigma_{\QQ}$ respectively. Then we have
$$
\KL{\PP}{\QQ} = {\frac 1 2} \ln \left( {\frac {\sigma_{\QQ}^2}
{\sigma_{\PP}^2}}\right) + {\frac {(\mu_{\PP} - \mu_{\QQ})^2 +
\sigma_{\PP}^2 - \sigma_{\QQ}^2} {2 \sigma_{\QQ}^2}}.
$$
\end{fact}

An easy consequence is the following bound on the KL divergence
between two Gaussians:
\begin{corollary} \label{cor:ubklgauss}
Let $\PP, \QQ$ be one-dimensional Gaussians as above and suppose
that $|\mu_{\PP} - \mu_{\QQ}| \leq \epsmeans$, $|\sigma_{\PP}^2 -
\sigma_{\QQ}^2|< \epsvars$, and $\sigma_{\PP}^2 \geq \sigmamin^2.$
Then
$$
\KL{\PP}{\QQ} \leq {\frac {\epsvars} {2 \sigmamin^2}} + {\frac
{\epsmeans^2 + \epsvars} {2(\sigmamin^2 - \epsvars)}}.
$$
\end{corollary}
\begin{proof}
We have
$$
{\frac {\sigma_{\QQ}^2} {\sigma_{\PP}^2}} \leq {\frac {\sigmamin^2 +
\epsvars}{\sigmamin^2}} = 1 + {\frac  {\epsvars}{\sigmamin^2}}
$$
which implies
$$
{\frac 1 2} \ln \left( {\frac {\sigma_{\QQ}^2}
{\sigma_{\PP}^2}}\right) \leq {\frac {\epsvars} {2 \sigmamin^2}}.$$
The bound easily follows observing that $\sigma_{\QQ}^2 \geq
\sigmamin^2 - \epsvars$.
\end{proof}

\begin{proposition} \label{prop:nKL}
Suppose $\PP_1, \dots, \PP_n$ and $\QQ_1, \dots, \QQ_n$ are
distributions satisfying $\KL{\PP_i}{\QQ_i} \leq \eps_i$ for all
$i$.  Then  $\KL{\PP_1 \times \cdots \times \PP_n}{\QQ_1 \times
\cdots \times \QQ_n} \leq \sum_{i=1}^n \eps_i$.
\end{proposition}
\begin{proof}
We prove the case $n = 2$:
\begin{eqnarray*}
\KL{\PP_1 \times \PP_2}{\QQ_1 \times \QQ_2} &= & \iint \PP_1(x)
\PP_2(y) \ln \frac{\PP_1(x)\PP_2(y)}{\QQ_1(x)\QQ_2(y)} dxdy \\
& = & \iint \PP_1(x) \PP_2(y) \ln \frac{\PP_1(x)}{\QQ_1(x)} dxdy +
\iint \PP_1(x) \PP_2(y) \ln \frac{\PP_2(y)}{\QQ_2(y)} dxdy \\
& = & \int \PP_2(y) \KL{\PP_1}{\QQ_1} dy + \int \PP_1(x)
\KL{\PP_2}{\QQ_2} dx\\
& \leq & \eps_1 + \eps_2. \vspace{-.2in}
\end{eqnarray*}
The general case follows by induction.
\end{proof}



\section{Truncated versus untruncated mixtures of Gaussians}
\label{ap:trunc}


\begin{definition} Let $\X$ be a distribution over $\R^n.$
The {\em $\M$-truncated version of $\X$} is the distribution $\X_\M$
obtained by restricting the support of $\X$ to be $[-\M,\M]^n$ and
scaling so that $\X_\M$ is a distribution.  More precisely, for $x
\in \R^n$ we have
\[
\X_\M(x) = \left\{
\begin{array}{ll}
0 & \text{if $\|x\|_\infty > \M$,} \\
c \X(x) & \text{if $\|x\|_\infty \leq \M$}
\end{array}
\right.
\]
where $c=1/\left(\int_{x \in [-\M,\M]^n} \X(x)\right)$ is chosen so
that $\int \X_\M(x) = 1.$
\end{definition}

In this section we prove Proposition \ref{prop:trunc}:

\medskip

\noindent {\bf Proposition \ref{prop:trunc}} {\em Let $\PP$ and
$\QQ$ be any mixtures of $n$-dimensional  Gaussians. Let $\PP_\M$
denote the $\M$-truncated version of $\PP$. For some
$\M=\poly(nL/\eps)$ we have $|\KL{\PP_M}{\QQ} - \KL{\PP}{\QQ}| \leq
4\eps + 2\eps \cdot \KL{\PP}{\QQ}$. }

\begin{proof}
We will take $M=M(\theta)$ (recall Appendix~\ref{sec:notconvgauss}).
As we go through the proof various conditions will be set on
$\theta$.  At the end of the proof we will see that we can take
$\theta=\poly(\eps/nL)$ and obtain the desired bound on
$|\KL{\PP_M}{\QQ} - \KL{\PP}{\QQ}|$ and satisfy all the conditions
on $\theta$.  This proves the theorem.

We have that $\PP_\M(x)$ satisfies
\[
\PP_\M(x) = \left\{
\begin{array}{ll}
(1+\delta) \PP(x) & \text{if $x \in [-\M,\M]^n$,} \\
0 & \text{if $x \not \in [-\M,\M]^n$,}
\end{array}
\right.
\]
where $\delta > 0$ is chosen so that $\frac{1}{1+\delta} = \int_{x
\in [-\M,\M]^n} \PP(x)$.  Using the definition of $\M$ we have
\[
\int_{x \not \in [-\M,\M]^n} \PP(x) = \Pr_\PP[x \not \in [-\M,\M]^n]
\leq \sum_{j=1}^n \Pr_\PP[|x_j| \geq \M] \leq n\theta \leq \eps
\]
where we have used the fact that $\theta \leq \eps/n$ (this is our
first condition on  $\theta$). Consequently we have ${\frac 1
{1+\delta}} \geq 1 - \eps$, so $\delta \leq 2\eps$.

We have
\begin{eqnarray*}
&& \! \! \! \! \! \! \! \! \! \! \! \! \! \! \! \! \! \! \! \!
\left| \KL{\PP_\M}{\QQ} - \KL{\PP}{\QQ} \right| \\
& = & \left| \int_{x \in [-\M,\M]^n} (1+\delta) \PP(x) \ln
\frac{(1+\delta)\PP(x)}{\QQ(x)} - \int_{x \in \R^n} \PP(x) \ln
\frac{\PP(x)}{\QQ(x)}\right|\\
& = & \left| (1+\delta) \ln(1+\delta)\int_{x \in [-\M,\M]^n} \PP(x)
+ \delta \int_{x \in [-\M,\M]^n} \PP(x) \ln \frac{\PP(x)}{\QQ(x)} -
\int_{x
\not \in [-\M,\M]^n} \PP(x) \ln \frac{\PP(x)}{\QQ(x)}\right| \\
& \leq & (1+\delta)\ln(1+\delta) + \delta \left| \int_{x \in
[-\M,\M]^n} \PP(x) \ln \frac{\PP(x)}{\QQ(x)} \right| + \left|\int_{x
\not
\in [-\M,\M]^n} \PP(x) \ln \frac{\PP(x)}{\QQ(x)}\right| \\
& = & \delta(1+\delta) + \delta |R| + |S|,
\end{eqnarray*}
where $R := \int_{x \in [-\M,\M]^n} \PP(x) \ln
\frac{\PP(x)}{\QQ(x)}$ and $S := \int_{x \not \in [-\M,\M]^n} \PP(x)
\ln \frac{\PP(x)}{\QQ(x)}$. For succinctness let $\kappa$ denote
$\KL{\PP}{\QQ}.$ Note that we have $\kappa = R + S$.

Suppose we show that $|S| \leq \eps$.  Then since $\kappa = R + S$,
we must have $|R| \leq \kappa + \eps$, and hence $|\KL{\PP_\M}{\QQ}
- \kappa| \leq \delta(1+\delta) + \delta(\kappa + \eps) + \eps \leq
4\eps + 2\eps\kappa$ (using $\delta \leq 2\eps$), as desired.  Thus
we can complete the proof by showing $|S| \leq \eps$.

Let us analyze the integrand of $S$.  Decompose $\PP$ into its
mixture components, i.e. $\PP(x) = \sum_{i=1}^k \pi^i \PP^i(x)$,
where $\PP^1, \dots, \PP^k$ are $n$-dimensional  Gaussians. Hence
\[
S = \sum_{i=1}^k \pi^i \int_{x \not \in [-\M,\M]^k} \PP^i(x) \ln
\frac{\PP(x)}{\QQ(x)}.
\]
We will show that for each $i$ we have $|\int_{x \not \in
[-\M,\M]^k} \PP^i(x) \ln \frac{\PP(x)}{\QQ(x)}| \leq \eps.$  It then
follows that $|S| \leq \eps$ since $|S|$ is upper bounded by a
convex combination of these quantities.

Let us now analyze the quantity $\ln \frac{\PP(x)}{\QQ(x)}$.  We
will show that for any $x\notin [-\M,\M]^k$, neither $\PP(x)$ nor
$\QQ(x)$ can be either ``too small'' or ``too large'' as a function
of $||x||_2^2$; hence $|\ln \frac{\PP(x)}{\QQ(x)}|$ will be of
moderate size. We will prove this for $\PP(x)$ using the fact that
it is a mixture of $n$-dimensional \bdd Gaussians; since this is
also true of $\QQ(x)$, the same bound will hold for it.

We will show that for all $i=1,\dots,k$ and all $x \in \R^n$ we have
$\PP^i(x) \in [t(x),T]$ where $T$ is a quantity and $t(x)$ is a
function that will both be defined below. Since $\PP(x) =
\sum_{i=1}^k \pi^i \PP^i(x)$ is a convex combination of the
$\PP^i(x)$'s, the same bound will hold for $\PP(x).$ Fix any $i$ and
consider the Gaussian $\PP^i.$ Since this Gaussian is axis-aligned,
we have $\PP^i(x) = \prod_{j=1}^n \phi_{\mu_j, \sigma_j^2}(x_j)$ for
some pairs $(\mu_1,\sigma_1^2), \dots, (\mu_n,\sigma_n^2)$
satisfying $|\mu_j| \leq \mumax$, $\sigma_j^2 \in [\sigmamin^2,
\sigmamax^2]$. (Here $\phi_{\mu,\sigma^2}(x)$ is the usual pdf
$\phi_{\mu,\sigma^2}(x) = {\frac 1 {\sqrt{2 \pi}\sigma}} \exp
\left({ \frac {-(x-\mu)^2}{2\sigma^2}}\right)$ for a one-dimensional
Gaussian.) It is easy to see that for any $x_j$,
\[
\frac{1}{\sqrt{2\pi}\sigmamax} \exp\left(-\frac{x_j^2}{\sigmamin^2}
- \frac{\mumax^2}{\sigmamin^2}\right) \leq \phi_{\mu_j,
\sigma_j^2}(x_j) \leq \frac{1}{\sqrt{2\pi}\sigmamin}.
\]
Hence for all $x \in \R^n$ we have
\begin{equation} \label{eqn:blah}
t(x) :=
\left(\frac{\exp(-\mumax^2/\sigmamin^2)}{\sqrt{2\pi}\sigmamax}\right)^n
\exp\left(-\frac{||x||_2^2}{\sigmamin^2}\right)\leq \PP^i(x) \leq
\left(\frac{1}{\sqrt{2\pi}\sigmamin}\right)^n =: T
\end{equation}
for all $i$, and so~(\ref{eqn:blah}) holds true for $\PP(x)$ as
well. As stated earlier, the same argument also shows
that~(\ref{eqn:blah}) holds for $\QQ(x)$.  We conclude that for any
$x$,
\begin{eqnarray*}
\left|\ln\frac{\PP(x)}{\QQ(x)}\right| & \leq & |\ln t(x)| + |\ln T| \\
& = & \left| -n \frac{\mumax^2}{\sigmamin^2} - n \ln (\sqrt{2\pi}
\sigmamax) - \frac{||x||_2^2}{\sigmamin^2}\right| + n
\ln(1/\sqrt{2\pi}\sigmamin) \\
& \leq & O\left(n
\frac{\mumax^2}{\sigmamin^2}\ln\frac{\sigmamax}{\sigmamin}
||x||_2^2\right).
\end{eqnarray*}

Recall that we want to show $|\int_{x \not \in [-\M,\M]^n} \PP^i(x)
\ln \frac{\PP(x)}{\QQ(x)}| \leq \eps$. It clearly suffices to show
that $\int_{x \not \in [-\M,\M]^n} \PP^i(x) |\ln
\frac{\PP(x)}{\QQ(x)}| \leq \eps$. By the above it suffices to show
\[
O\left(n
\frac{\mumax^2}{\sigmamin^2}\ln\frac{\sigmamax}{\sigmamin}\right)
\int_{x \not \in [-\M,\M]^n} \PP^i(x) ||x||_2^2 \leq \eps.  \] We
have
\begin{eqnarray}
\int_{x \not \in [-\M,\M]^n} \PP^i(x) ||x||_2^2 = \sum_{j=1}^n
\int_{x \not \in [-\M,\M]^n} \PP^i(x) x_j^2 \label{eq:theone}
\end{eqnarray}
Fix $j$; we now bound $\int_{x \not \in [-\M,\M]^n} \PP^i(x) x_j^2$.
Recall that $\PP^i(x) = \PP^i_1(x_1) \cdots \PP^i_n(x_n).$ We have
\begin{eqnarray*}
\int_{x \not \in [-\M,\M]^n} \PP^i(x) x_j^2 &\leq&  \sum_{\ell =
1}^n \int_{x \in \R^n: |x_\ell| > \M}
\PP^i(x) x_j^2 \nonumber \\
& = & \int_{x \in \R^n : |x_j| >\M} \PP^i(x) x_j^2 + \sum_{\ell \neq
j} \int_{x \in \R^n: |x_\ell| > \M} \PP^i(x)x_j^2 \label{eq:jl}
\end{eqnarray*}
For the first integral of (\ref{eq:jl}) above we have
\begin{eqnarray}
\int_{x \in \R^n : |x_j| >\M} \PP^i(x) x_j^2  = \left(\prod_{\ell
\neq j} \left[\int_{x_\ell \in \R}\PP^i_{\ell}(x_\ell)
dx_{\ell}\right]\right) \cdot \int_{|x_j| > \M} \PP^i_j(x_j)x_j^2
dx_j &=&
\int_{|x_j| > \M} \PP^i_j(x_j)x_j^2 dx_j \nonumber \\
&\leq& \theta \label{eq:th1}
\end{eqnarray}
where the inequality is by the definition of $\M.$ For the second
term of (\ref{eq:jl}) above we have
\begin{eqnarray}
\sum_{\ell \neq j} \int_{x \in \R^n: |x_\ell| > \M} \PP^i(x)x_j^2
&=& \sum_{\ell \neq j} \left[ \left(\int_{|x_\ell| >
\M}\PP^i_\ell(x_\ell) dx_\ell\right) \ignore{\left(\prod_{\ell' \neq
\ell} \left[\int_{x_{\ell'} \in \R}\PP^i_{\ell'} (x_{\ell'})
dx_{\ell'}\right]\right)} \left(\int_{x_j \in \R}\PP^i_j(x_j)x_j^2
dx_j\right)\right] \label{eq:um}
\end{eqnarray}
where we have used the fact that for any $\ell'$ which is neither
$\ell$ nor $j$ we have
$$\int_{x_{\ell'} \in \R} \PP^i_{\ell'}(x_{\ell'}) dx_{\ell'} = 1.
$$
Again using the definition of $\M$ to bound the integral over
variable $x_\ell$ in (\ref{eq:um}) above by $\theta$, we have that
(\ref{eq:um}) is at most
\begin{eqnarray}
(n-1)\theta \int_{x_j \in \R}\PP^i_j(x_j)x_j^2 dx_j =(n-1)\theta
\E_{\PP^i_j}[x^2] &=&
(n-1)\theta \left(\Var_{\PP^i_j}[x] + \E_{\PP^i_j}[x]^2\right)\nonumber \\
&=&
(n-1)\theta ((\sigmaij)^2 + (\muij)^2) \nonumber \\
&\leq& (n-1)\theta (\sigmamax^2 + \mumax^2) \label{eq:bd}
\end{eqnarray}
where the inequality holds since $\PP^i_j$ is a one-dimensional \bdd
Gaussian.

Putting all the pieces together, we find that (\ref{eq:theone}) is
at most
$$
n[\theta + (n-1) \theta(\sigmamax^2 + \mumax^2)] \leq n^2
\theta(\sigmamax^2 + \mumax^2)$$ It follows that $|S| \leq n^2
\theta(\sigmamax^2 + \mumax^2) \cdot O(n^2
\frac{\mumax^2}{\sigmamin^2}\ln\frac{\sigmamax}{\sigmamin})$.  We
can take $\theta=\poly(\eps/nL)$ and have this quantity be at most
$\eps.$
\end{proof}

\end{document}